\documentclass[pmlr]{jmlr}
\usepackage{amsmath,amssymb,graphicx,url}
\jmlrvolume{329}
\jmlryear{2026}
\jmlrworkshop{Conformal and Probabilistic Prediction with Applications}
\jmlrproceedings{PMLR}{Proceedings of Machine Learning Research}

\title[Probabilistic Object Detection with Conformal Prediction]{Probabilistic Object Detection with Conformal Prediction}
 \author{\Name{Christopher Ries} \Email{uliaz@student.kit.edu}\\
  \Name{Moussa Kassem~Sbeyti} \Email{moussa.sbeyti@kit.edu}\\
  \Name{Nicolas Bianco} \Email{nicolas.bianco@kit.edu}\\
  \Name{Nadja Klein} \Email{nadja.klein@kit.edu}\\
  \addr Karlsruhe Institute of Technology (KIT), Karlsruhe, Germany}

\editor{Ernst Ahlberg, Ulf Johansson, Henrik Boström, Alberto Carlevaro, Johan Hallberg Szabadváry and Lars Carlsson}

\usepackage{longtable}

\usepackage{mathtools}
\usepackage{bbm}
\usepackage{tablefootnote}
\usepackage{booktabs}
\usepackage{acronym}

\acrodef{CP}{Conformal Prediction}
\acrodef{DNN}{Deep Neural Networks}
\acrodef{PI}{Prediction Interval}
\acrodef{UQ}{Uncertainty Quantification}
\acrodef{RAPS}{Regularized Adaptive Prediction Sets}
\acrodef{APS}{Adaptive Prediction Sets}
\acrodef{LAS}{Least Ambigious Set-Valued}
\acrodef{IoU}{Intersection over Union}
\acrodef{MC}{Monte Carlo}
\acrodef{IR}{isotonic regression}
\acrodef{PC}{per-class}
\acrodef{PCo}{per-coordinate}
\theorembodyfont{\upshape}
\theoremheaderfont{\scshape}
\theorempostheader{:}
\theoremsep{\newline}

\begin{document}

\maketitle

\begin{abstract}
Conformal Prediction (CP) is a distribution-free method for constructing prediction sets with marginal finite-sample coverage guarantees, making it a suitable framework for reliable uncertainty quantification in safety-critical object detection. However, object detection introduces structured multi-output predictions, complicating the application of classical CP theory developed for single outputs. In addition, standard, unscaled CP produces fixed-width prediction intervals across inputs, leading to unnecessary width for low-uncertainty predictions. While scaled CP addresses this by adapting the interval width to an input-dependent uncertainty estimate, prior work has neither systematically compared unscaled and scaled CP for multi-class object detection, nor integrated CP with a complementary uncertainty quantification method in this setting.
We fill this gap by: (i) applying CP coordinate-wise to bounding box corners with a Bonferroni correction for box-level guarantees; (ii) scaling the resulting intervals using per-prediction aleatoric uncertainty estimates derived from a probabilistic object detector trained with loss attenuation, evaluated in uncalibrated and two calibrated variants; (iii) extending to a two-step pipeline that constructs prediction sets for the class using \ac{RAPS} and conditions the conformalized bounding boxes on the predicted class set.
Across three autonomous driving datasets (KITTI, BDD, CODA), including a cross-domain setting under distribution shift, scaled CP consistently improves interval sharpness over unscaled CP, achieving up to 19\% higher IoU and 39\% lower interval scores, without sacrificing coverage. Class-wise calibration further improves coverage for both variants with a negligible effect on sharpness. Together, these improvements yield more actionable uncertainty estimates for real-time, real-world object detection. Code is available at \url{https://github.com/mos-ks/OD-CP}.

\end{abstract}

\section{Introduction}
\label{sec:intro}
Deploying object detectors in safety-critical applications such as autonomous driving requires not only accurate predictions but also reliable associated uncertainty estimates. Without the latter, a model that is wrong with high confidence poses a direct risk. However, state-of-the-art object detectors rely on deterministic neural networks that regress bounding boxes and classify objects with corresponding sigmoid-based scores without any uncertainty estimation. While several probabilistic object detectors have been proposed~\citep{ChoChunKim2019, FenHarWas2022}, the underlying \ac{UQ} methods are typically model-specific as well as computationally expensive, thereby limiting their broad applicability~\citep{GawTasAli2023}. \ac{CP} addresses this limitation: it is a post-hoc, model-agnostic, and distribution-free framework. Given a pre-defined miscoverage level $\alpha \in [0,1]$, \ac{CP} constructs prediction intervals for regression or prediction sets for classification that contain the true value with probability of at least $1-\alpha$~\citep{PapProVov2002,VovGamSha2005,Vov2013}. This guarantee holds in finite samples without assumptions on the underlying model, relying solely on the exchangeability of the calibration and test data. However, CP in its original form has been designed for single outputs and is thus not directly applicable to the multi-output predictions of object detectors. 

Standard (unscaled) \ac{CP} produces prediction intervals of constant width across inputs. Therefore, it ignores the varying difficulty of individual predictions. Scaled \ac{CP}, instead, addresses this by adapting the interval width to an input-dependent uncertainty estimate. In object detection, a natural candidate for this estimate is aleatoric uncertainty, which captures irreducible noise inherent in the data and is thus well-suited to characterize the difficulty of individual predictions at inference time~\citep{KasSbeyKar2023,KasSbeyKar2024}. Loss attenuation~\citep{KenGal2017} provides a computationally efficient, sampling-free means to estimate this uncertainty by learning a per-prediction variance alongside the regression target. However, the behaviour of aleatoric uncertainty-scaled \ac{CP} in the context of multi-class object detection remains poorly understood. In particular, no prior work has systematically studied whether scaling \ac{CP} with such estimates improves over the unscaled baseline, nor how uncertainty calibration affects this scaling.

In this work, we build on an EfficientDet-based probabilistic object detector trained with loss attenuation~\citep{KasSbeyKar2023}, which provides per-prediction aleatoric uncertainty estimates. We use these estimates as scaling factors for scaled \ac{CP}, evaluated in three forms: raw (uncalibrated) and two 
calibrated variants based on isotonic regression~\citep{BarBru1972,KulFenErm2018}, applied either globally or per-coordinate and per-class. We compare these against unscaled \ac{CP} across class-agnostic and class-wise settings on three autonomous driving datasets. The three datasets cover both in-domain evaluation and a cross-domain setting, representing a realistic distribution shift scenario. Furthermore, following~\citet{TimStrSak2025}, we evaluate a two-step pipeline that applies \ac{CP} to both the regression and classification heads of the object detector, using \ac{RAPS}~\citep{AngBatMal2022} to produce class-conditional prediction sets that condition the conformalized bounding boxes.

\begin{figure*}[t!]
    \centering
    \includegraphics[width=1\linewidth]{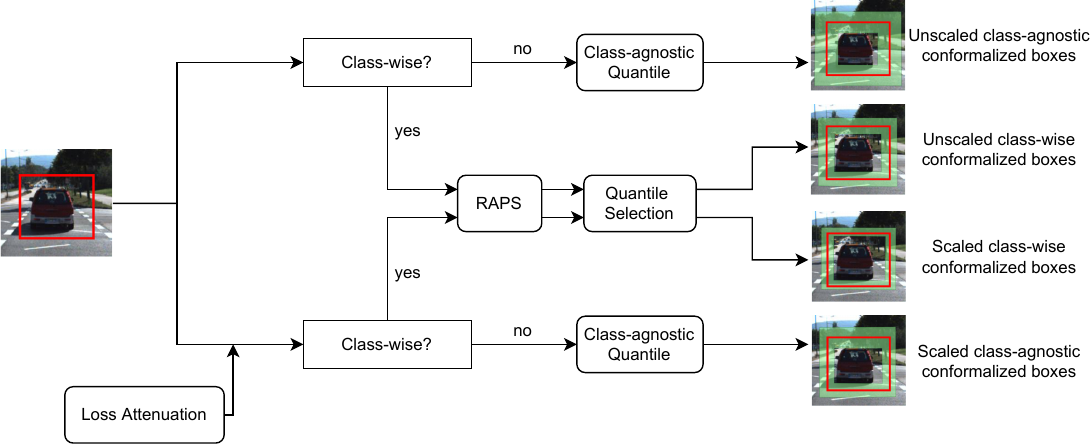}
    \caption{Overview of the pipeline. The upper branch applies unscaled \ac{CP} and the lower branch applies scaled \ac{CP}, where the prediction intervals are scaled by aleatoric uncertainty estimates derived via loss attenuation. In both branches, the class-wise path additionally computes prediction sets for the classification head using RAPS, followed by class-conditional quantile selection.}
    \label{fig:our_pipeline}
\end{figure*}

Our pipeline is shown in Figure~\ref{fig:our_pipeline}. The contributions of this work are:
\begin{itemize}
    \item[(i)] We provide a systematic comparison of unscaled and scaled \ac{CP} for multi-class object detection, evaluated not only on coverage but also on interval sharpness metrics including \ac{IoU} and interval score, across both in-domain and cross-domain settings under distribution shift.
    \item[(ii)] We study the effect of uncertainty calibration on scaled \ac{CP}, comparing raw loss attenuation estimates against two calibrated variants.
    \item[(iii)] We evaluate class-wise \ac{CP} for bounding box regression and demonstrate its effect on coverage and sharpness relative to the class-agnostic setting.
    \item[(iv)] We evaluate a two-step pipeline that jointly conformalize the classification and regression heads.
\end{itemize}

In summary, we demonstrate that scaled \ac{CP} consistently achieves sharper intervals than unscaled \ac{CP} without sacrificing the validity of the coverage guarantee, and that this finding holds across all datasets and experimental settings considered.

\section{Related Work}
\label{sec:related_work}
\textbf{Uncertainty Quantification in Deep Learning.}
Uncertainty in deep learning models can be decomposed into epistemic uncertainty, reflecting model ignorance that can in principle be reduced with more data, and aleatoric uncertainty, capturing irreducible noise inherent in the observations themselves~\citep{KenGal2017}. Methods for estimating these uncertainties broadly fall into two categories. Distributional approaches modify the model or training procedure to produce uncertainty estimates directly. Common examples include deep ensembles~\citep{LakPriBlun2017} and \ac{MC} dropout~\citep{GalGha2016} for epistemic uncertainty, and loss attenuation~\citep{KenGal2017} for aleatoric uncertainty in regression. Sampling-based methods such as ensembles and \ac{MC} dropout are computationally expensive, requiring either multiple training runs or multiple forward passes at inference~\citep{GawTasAli2023}. Methods such as \ac{MC} dropout additionally require architectural modifications, limiting their applicability. Post-hoc approaches, by contrast, are applied to a fixed pre-trained model and make no assumptions about the underlying architecture, making them implementation-agnostic. Calibration methods such as isotonic regression~\citep{KulFenErm2018} and conformal prediction~\citep{VovGamSha2005} fall into this category. 

\textbf{Uncertainty Quantification in Object Detection.}
Object detection introduces specific challenges for uncertainty quantification compared to standard classification or regression tasks. Predictions are structured and multi-output: a detector produces both a bounding box and a class label for each detected object, each carrying its own source of uncertainty~\citep{ChoElelee2021}. The localization output is a four-dimensional continuous variable, and object detectors operate on cluttered scenes with varying object scales, occlusion levels, and lighting conditions, all of which affect prediction difficulty in an input-dependent manner. General \ac{UQ} methods such as \ac{MC} dropout and deep ensembles have been applied to object detection~\citep{HarSmaWas2020,LyuGutRajBek2020}, but their computational cost and architectural requirements limit their practical use in real-time detection pipelines~\citep{GawTasAli2023}. Moreover, epistemic uncertainty reflects model ignorance that diminishes with more data and a better model, and is therefore less informative about the difficulty of individual predictions at inference time. 
Aleatoric uncertainty, by contrast, is tied to the irreducible noise in each input and remains informative at inference time even for a well-trained model. Crucially, \citet{KasSbeyKar2023} demonstrate that aleatoric uncertainty estimated via loss attenuation correlates with task-relevant error sources such as occlusion, object distance, and image quality in object detection, making it a well-motivated scaling signal for input-adaptive prediction intervals. Probabilistic object detectors capture this by modelling the bounding box output as a Gaussian distribution $N(\mu, \sigma^2)$, where $\sigma^2$ represents the per-coordinate aleatoric uncertainty~\citep{ChoChunKim2019}. Loss attenuation~\citep{KasSbeyKar2023} achieves this by learning the variance alongside the regression target through a heteroscedastic negative log-likelihood objective, without requiring multiple forward passes or architectural changes beyond the output head. However, the raw estimates produced by loss attenuation are typically miscalibrated~\citep{KasSbeyKar2023}, motivating post-hoc calibration via isotonic regression before being used as scaling factors for \ac{CP}.

\textbf{Conformal Prediction in Deep Learning.}
\ac{CP}~\citep{PapProVov2002,VovGamSha2005,Vov2013} has emerged as a post-hoc and model-agnostic approach to uncertainty quantification, offering finite-sample marginal coverage guarantees under the assumption of exchangeability of the calibration and test data. Unlike the distributional and sampling-based methods discussed above, \ac{CP} requires only a single model evaluation and makes no assumptions about the underlying architecture, making it applicable to any black-box predictor. For regression, \ac{CP} constructs prediction intervals around the model output; for classification, it constructs prediction sets containing the true class with a user-specified probability. Several algorithms have been proposed for the classification setting, including \ac{APS}~\citep{RomSesCan2020} and \ac{RAPS}~\citep{AngBatMal2022}, which build adaptive prediction sets by accumulating sorted class probabilities until a calibrated threshold is reached. \ac{CP} in deep learning has demonstrated strong potential in safety-critical domains including medical imaging~\citep{LuLemCha2022,VazFac2022,GhoWooMen2025}, autonomous driving~\citep{deGAdaAle2022,TimStrSak2025}, and railway signaling~\citep{AndFelGra2023}. A key requirement is exchangeability, which requires the joint distribution of calibration and test data to be invariant to finite permutations~\citep{VovGamSha2005,Vov2013}. This assumption may be violated when calibration and deployment conditions differ, for example under domain shift~\citep{TibFoyBarCan2019}, which we explicitly investigate in our cross-domain evaluation.

\textbf{Conformal Prediction in Object Detection.}
Applying \ac{CP} to object detection is non-trivial for two reasons. First, bounding box regression is a multi-output problem: \ac{CP} must be applied to each of the four box coordinates independently, requiring a multiple testing correction to obtain joint box-level coverage guarantees. The standard choice is the Bonferroni correction~\citep{BlAlt1995}, which allocates a miscoverage of $\alpha/4$ to each coordinate, ensuring an overall miscoverage no greater than 
$\alpha$. While conservative~\citep{deGAdaAle2022,TimStrSak2025}, it does not require assumptions about the dependence structure of the coordinate-wise errors. Alternatives that exploit the dependence structure between coordinates, such as copula-based corrections~\citep{MukMesRou2024}, have been proposed and can yield tighter intervals. However, they require estimating the joint dependence structure from the calibration data, adding implementation complexity and potentially increasing sensitivity to distribution shift. Hence, in this paper, we adopt the Bonferroni approach for simplicity and comparability with prior work~\citep{deGAdaAle2022,TimStrSak2025}. Beyond the choice of correction, previous work differ in which heads of the detector \ac{CP} is applied to and whether fixed or adaptive interval widths are used. \Citet{deGAdaAle2022} apply unscaled \ac{CP} exclusively to the localization head, comparing coordinate-wise, box-wise, and image-wise variants, and conclude that coordinate-wise and box-wise \ac{CP} are better suited for pedestrian localization. However, they do not address the classification head or scaled \ac{CP}. The work most closely related to ours is by~\citet{TimStrSak2025}, who extend \ac{CP} to both the classification and regression heads through a two-step pipeline: prediction sets for the class are first computed, and the conformalized bounding boxes are then conditioned on the predicted class. They further introduce a scaled variant based on a conformal ensemble. While their approach demonstrates the value of adaptive interval width, it relies on an ensemble, making it computationally expensive. In contrast, our work evaluates scaling based on aleatoric uncertainty derived from loss attenuation, which is sampling-free and requires only a single forward pass. Furthermore, unlike~\citet{TimStrSak2025}, we systematically compare unscaled and scaled \ac{CP} across multiple datasets and both calibrated and uncalibrated uncertainty estimates, evaluating performance on coverage and sharpness metrics and explicitly testing robustness under distribution shift.

\section{Method}
\label{sec:method}
In this section, we first provide the necessary background on \ac{CP} for regression and classification, and then describe how \ac{CP} is applied to object detection to obtain prediction intervals for bounding boxes as well as prediction sets for object classes, forming the basis of our experimental evaluation.

\subsection{Conformal Prediction} 
Let $Y\in\mathcal{Y}\subseteq\mathbb{R}$ and $X\in\mathcal{X}\subseteq\mathbb{R}^p$ be an outcome and $p$-dimensional input variable, respectively. Let $(X_i,Y_i)\sim P_{XY}$ be independent and identically distributed pairs of input vectors and output $i=1,\ldots n$ with unknown on $\mathcal{X} \times \mathcal{Y}$. \ac{CP} aims to provide a prediction interval (for regression) or set (for classification) $\mathcal{C}(X_{n+1})$ such that for some new input and output pair $(X_{n+1},Y_{n+1})\sim P_{XY}$  for any fixed miscoverage level $\alpha\in[0,1]$:
\begin{equation}
 \mathbb{P}(Y_{n+1} \in \mathcal{C}(X_{n+1}))\geq 1-\alpha,
    \label{eq:marginal_coverage}
\end{equation}
where the quantity $1-\alpha$ is also called \emph{marginal coverage}, and the probability is taken with respect to $P_{XY}$. 

According to how $\mathcal{C}(X_{n+1})$ is computed, \ac{CP} can be categorized into \emph{full}~\citep{VovGamSha2005} and \emph{split}~\citep{VovGamSha2005,Vov2013}. The first one requires to fit a model for each possible one-hold-out split of the dataset, therefore being accurate but computationally demanding. On the other hand, split \ac{CP} considers a single split into a calibration and evaluation set, often leading to larger prediction interval/set~\citep{VovGamSha2005,Vov2015} but at a reduced computational cost~\citep{AngBat2021}. 

In this work, we consider split \ac{CP}. Assume that we observe $n+m$ samples $\mathcal{D}=\{(y_i,x_i)\}_{i=1}^{n+m}$ that have not been used during training and randomly split $\mathcal{D}$ into a calibration set $\mathcal{D}_{\text{cal}}$ of size $n$ and an evaluation set $\mathcal{D}_{\text{eval}}$ of size $m$. Moreover, we denote with $\hat{f}:\mathcal{X}\rightarrow \mathcal{Y}$ a prediction function and with $\hat{y}_i=\hat{f}(x_i)$ the prediction of the outcome at input $x_i$. Define a score function $s(\hat{y},y)$ that returns higher values for worse predictions $\hat{y}$ of $y$, and $s(\hat{y},y)=0$ if and only if $\hat{y}=y$. The score function is also called \emph{conformity score}. Its choice determines the width and adaptivity of the resulting prediction intervals or sets (scaled vs.~ unscaled), and we describe the relevant options for regression and classification separately below.

\textbf{Regression.}
Our aim is to create a prediction interval $[l(x), u(x)]$ based on:
\begin{align}
    \text{(Unscaled)}\quad s(\hat{y},y) &= |\hat{f}(x) - y|, \label{eq:score_func_residuals}\\ 
    \text{(Scaled)}\quad s(\hat{y},y) &= \frac{|\hat{f}(x) - y|}{\hat{\sigma}(x)}. \label{eq:score_func_residuals_scaled}
\end{align}
The score function \eqref{eq:score_func_residuals} corresponds to the absolute value of the regression residuals, while \eqref{eq:score_func_residuals_scaled} rescales the residuals by input-dependent uncertainty estimate $\hat{\sigma}(x)>0$~\citep{LeiGSe2018}. The prediction intervals are constructed as:
\begin{align*}
    &\text{(Unscaled)} &&\widehat{\mathcal{C}}(x) = [\hat{f}(x) - \hat{q}, \hat{f}(x) + \hat{q}]\\ 
    &\text{(Scaled)} &&\widehat{\mathcal{C}}(x) = [\hat{f}(x) - \hat{q} \cdot \hat{\sigma}(x), \hat{f}(x) + \hat{q}\cdot \hat{\sigma}(x)],
\end{align*}
for the unscaled and scaled \ac{CP}, respectively, where $\hat{q}$ is the $\lceil(n+1)(1-\alpha)\rceil/n$-quantile of the respective score vector $s=(s(\hat{y}_1,y_1),\ldots,s(\hat{y}_n,y_n))^\top$ computed on the calibration set $\mathcal{D}_{\text{cal}}=\{(y_i,x_i)\}_{i=1}^n$. 

For both, unscaled and scaled CP, the marginal coverage defined in \eqref{eq:marginal_coverage} holds. Empirically, this means that for the samples in the evaluation set $\mathcal{D}_{\text{eval}}=\{(y_i,x_i)\}_{i=1}^m$ it holds that 
\[
\frac{1}{m}\sum_{i=1}^m\mathbb{I}\left(y_i \in \widehat{\mathcal{C}}(x_i)\right) \in \left(1-\alpha,1-\alpha + \frac{1}{n+1}\right),
\]
where $\mathbb{I}(\cdot)$ is the indicator function.

In this work, the learned variance $\hat{\sigma}(x)$ via loss attenuation serves as the input-dependent scaling factor in Equation~\eqref{eq:score_func_residuals_scaled}, directly linking aleatoric uncertainty to the width of the conformalized prediction intervals.

\textbf{Classification.}
Assume a multi-class classification problem with $K$ classes. In this context, the prediction $\hat{y}=\hat{f}(x)$ is a $K$-dimensional vector of per-class probabilities $\hat{y}=(\hat{y}_1,\ldots,\hat{y}_K)^\top$ for some input value $x$. Let $\pi(x)$ be the permutation such that the output probabilities $\hat{y}_{\pi(x)}$ are sorted from highest to lowest. The prediction set is then constructed by including the highest probability classes in order until a calibrated threshold is reached. The choice of score function determines how this threshold is computed; in this work we consider \ac{APS}~\citep{RomSesCan2020,AngBat2021} and \ac{RAPS}~\citep{AngBatMal2022}, which differ as follows:
\begin{align}
    \text{(APS)}\quad s(\hat{y},y) &= \sum_{j=1|\hat{y}_j \leq \hat{y}_\star}^k \hat{y}_{\pi_j(x)}, \\
        \text{(RAPS)}\quad s(\hat{y},y) &= \sum_{j=1|\hat{y}_j \leq \hat{y}_\star}^k \left(\hat{y}_{\pi_j(x)}+a\mathbb{I}(j>b)\right),
    \label{eq:score_func_class}
\end{align}
where $\hat{y}_\star$ is the probability of the true class $k_\star$. Note that RAPS score function is built to encourage small prediction sets and depends on two hyperparameters $(a,b)$. The role of these is explained in the work of~\cite{AngBatMal2022}. In this work, we use the hyperparameter suggested in the work of~\cite{AngBat2021}. During inference, the prediction set is defined as the collection of $k$ classes $\widehat{\mathcal{C}}(x) = \{\pi_1(x), \dots, \pi_k(x)\}$ such that:
\begin{equation}
    \begin{split}
        k = \sup \left\{
        \sum_{j=1}^{k'} \hat{f}(x)_{\pi_j(x)} < \hat{q}
        \right\} + 1,
    \end{split}
\end{equation}
where $\hat{q}$ is the $\lceil(n+1)(1-\alpha)\rceil/n$-quantile of the score vector $s=(s(\hat{y}_1,y_1),\ldots,s(\hat{y}_n,y_n))^\top$ computed on the calibration set $\mathcal{D}_{\text{cal}}=\{(y_i,x_i)\}_{i=1}^n$. This construction of the predictions set achieve the theoretical marginal coverage in \eqref{eq:marginal_coverage}, while empirically it holds that for the samples in the evaluation set $\mathcal{D}_{\text{eval}}=\{(y_i,x_i)\}_{i=1}^m$:
\[
\frac{1}{m}\sum_{i=1}^m\mathbb{I}\left(k_{i,\star} \in \widehat{\mathcal{C}}(x_i)\right) \in \left(1-\alpha,1-\alpha + \frac{1}{n+1}\right).
\]

RAPS also has the ability to enforce non-empty prediction sets~\citep{AngBat2021, AngBatMal2022}. This is achieved by first checking if the prediction set is empty, and if it is, adding the class with the largest softmax output to the prediction set.

\subsection{Applying Conformal Prediction to Object Detection} \label{sec:joint_cp}
Let $I \in \mathbb{R}^{H \times W \times D}$ be an input image $H$, $W$ and $D$ corresponding to height, width, and depth, respectively. For each image, a vector of coordinates  $(c_{x,0}, c_{y,0}, c_{x,1}, c_{y,1}) \in \mathbb{R}^4$ corresponding to the corners of the object present in the image is available together with the true label of the class $k \in\{1,\dots,K\}$ to which the object belongs to. The true bounding box is the boundary of the rectangle $A=[c_{x,0}, c_{x,1}]\times [c_{y,0}, c_{y,1}]$, which we denote by $\delta(A)$.

\ac{CP} in the context of object detection can be used to i) obtain prediction intervals around the bounding box (regression) or ii) obtain these jointly with a prediction set for the class of the detected object (regression and classification).

\textbf{Class-Agnostic Conformal Prediction: Regression.}
The first task requires to construct a prediction interval for the bounding box using a single set of conformity scores computed across all classes, regardless of the object category. In a class-agnostic setting and for a new input image $I$ with true bounding box $\delta(A)$, the \ac{CP} procedure should provide a predictive area $\mathcal{C}(I)= A^+\setminus A^-$ where:
\begin{align*}
    A^+ &= [a^+_{x,0},a^+_{x,1}]\times [a^+_{y,0},a^+_{y,1}]\\
    A^- &= [a^-_{x,0},a^-_{x,1}]\times [a^-_{y,0},a^-_{y,1}].
\end{align*}
such that $\mathbb{P}(\delta(A) \in \mathcal{C}(I)) = 1 - \alpha_{\text{bbox}}$, for $\alpha_{\text{bbox}}\in[0,1]$. Note that $A^+$ and $A^-$ can be defined in terms of the prediction intervals for each corner, where \begin{align*}
    a^+_{i,j}&=\mathcal{C}^+_{i,j}(I)=\max_i\mathcal{C}_{i,j}(I) \\
    a^-_{i,j}&=\mathcal{C}^-_{i,j}(I)=\min_i\mathcal{C}_{i,j}(I),
\end{align*}
for $i=x,y$ and $j=0,1$. Define the event that the coordinate $c_{i,j}$ is included in the prediction interval $\mathcal{C}_{i,j}(I)$ as $E_{i,j}(I)=\left\{c_{i,j} \in \mathcal{C}_{i,j}(I)\right\}$, and let $E(I)=\cap_{i\in\{x,y\},j\in\{0,1\}}E_{i,j}(I)$ be the event that each corner of the bounding box is included in the prediction interval simultaneously. Then,
\[
\mathbb{P}(\delta(A) \in \mathcal{C}(I))=\mathbb{P}\left(E(I)\right) = 1 - \alpha_{\text{bbox}},
\]
for $\alpha_{\text{bbox}}\in[0,1]$. However, since we apply \ac{CP} to each of the four corners individually, we can only obtain a lower bound on the aforementioned probability. Assume that we require the same marginal coverage $1-\alpha$ per each corner, applying Boole's inequality it holds that:
\begin{align}
        \mathbb{P}\left(E(I)\right) &=1-\mathbb{P}\left(\cup_{{i\in\{x,y\},j\in\{0,1\}}} E_{i,j}^c(I)\right)  \nonumber\\
        & \geq  1-\sum_{{i\in\{x,y\},j\in\{0,1\}}}\mathbb{P}\left(E_{i,j}^c(I)\right) \nonumber\\
        & = 1 - 4\alpha,
    \label{eq:class_agnostic_cp}
\end{align}
where $E^c_{i,j}(I)=\left\{c_{i,j} \not\in \mathcal{C}_{i,j}(I)\right\}$ is the complementary event of $E_{i,j}(I)$. It follows that choosing $\alpha=\alpha_{\text{bbox}}/4$ is a sufficient condition to ensure a lower bound probability of $1-\alpha_{\text{bbox}}$ on the coverage of the entire bounding box. This is the Bonferroni correction applied to the four bounding box coordinates~\citep{BlAlt1995}, and it does not require any assumptions about the dependence structure of the coordinate-wise errors.

\textbf{Class-Conditional Conformal Prediction: Regression.}
Equation~(\ref{eq:class_agnostic_cp}) only provides marginal coverage across all classes but not a class-conditional coverage. To achieve this, we use class-wise CP, where the calibration set is split in smaller sub-sets according to the ground truth class, where the latter is assumed to be known. The conformal scores and prediction set are computed per each class $k\in\{1,\ldots,K\}$:
\begin{equation}
        \mathbb{P}\left(E(I)\mid Y = k\right) \geq 1 - \alpha_{\text{bbox}}.
    \label{eq:class_wise_cp}
\end{equation}
The advantage of this approach is to adapt the width of prediction intervals to each class. However, in this setting, the calibration set cannot be selected at random. Instead, it must be constructed through stratified sampling to ensure that each class is represented by at least a minimum number of samples.

\textbf{Joint Conformal Prediction: Classification and Regression.}
While Equation~\eqref{eq:class_wise_cp} assumes that the class $k$ used for conditioning is known at inference time, this is not the case in practice. The true class must therefore be predicted, and its uncertainty quantified jointly with that of the bounding box, yielding a prediction set for the object class and a prediction interval for the bounding box coordinates simultaneously.

A naive solution applies CP with coverage $1-\alpha_{\text{bbox}}$ to each class $k\in\{1,\ldots,K\}$ separately and takes the widest prediction interval across all classes as a worst-case bound. While this guarantees coverage, it discards class information entirely and leads to overly large intervals.

The second alternative, following~\citet{TimStrSak2025}, first applies CP to the classification head to obtain a prediction set $\mathcal{C}_{\text{class}}(I)$ for the unknown true class. The bounding box quantile is then selected as the worst case over only the classes in $\mathcal{C}_{\text{class}}(I) \subset \{1,\ldots,K\}$ rather than all $K$ classes, yielding tighter intervals while maintaining coverage. The sharpness of the conformalized boxes therefore depends directly on the size of the predicted class set: smaller sets yield smaller worst-case quantiles and hence tighter boxes. This motivates a classification CP method that favours small prediction sets, namely RAPS, whose score function is explicitly regularized to that end. Under the assumption that CP for the class and bounding box are independent,~\citet{TimStrSak2025} show that this two-step procedure achieves the following joint coverage bound:
\begin{equation*}
    \begin{split}
            \mathbb{P}  \left(\{k \in \mathcal{C}_{\text{class}}(I)\} \land E(I) \mid Y = k\right) &= \mathbb{P}(k \in  \mathcal{C}_{\text{class}}(I) | Y=k) \mathbb{P}(E(I)\mid Y = k) \\
            &\geq (1-\alpha_{\text{class}}) (1-\alpha_{\text{bbox}}), 
    \end{split}
    \label{eq:cp_for_class_and_bbox_coverage}
\end{equation*}
where $\alpha_{\text{class}}$ and $\alpha_{\text{bbox}}$ are the miscoverage levels for the classification and bounding box regression steps, respectively. However, in object detection this independence assumption is unlikely to hold, since factors such as occlusion, object distance, and poor image quality tend to increase both classification and localization difficulty simultaneously, positively associating the two coverage events rather than leaving them independent. In this work, we relax the independence assumption, and replace it with a milder assumption of non-negative correlation. Let $\mathbb{I}_{\text{class}},\mathbb{I}_{\text{bbox}}\in\{0,1\}$ be the indicator functions of the events $\{k \in \mathcal{C}_{\text{class}}(I)\}$ and $E(I)$, respectively. Then, 
\begin{align*}
            \mbox{Cov}(\mathbb{I}_{\text{class}},\mathbb{I}_{\text{bbox}}\mid Y = k) &= \mathbb{E}(\mathbb{I}_{\text{class}}\mathbb{I}_{\text{bbox}}\mid Y = k) - \mathbb{E}(\mathbb{I}_{\text{class}}\mid Y = k)\mathbb{E}(\mathbb{I}_{\text{bbox}}\mid Y = k) \\
            &= \mathbb{P}\left(\{k \in \mathcal{C}_{\text{class}}(I)\} \land E(I) \mid Y = k\right) \\
            &\qquad - \mathbb{P}(k \in  \mathcal{C}_{\text{class}}(I) | Y=k) \mathbb{P}(E(I)\mid Y = k). 
\end{align*}
Rearranging terms and using classification and bounding box miscoverage levels we get:
\begin{align*}
            \mathbb{P}\left(\{k \in \mathcal{C}_{\text{class}}(I)\} \land E(I) \mid Y = k\right) &\geq (1-\alpha_{\text{class}}) (1-\alpha_{\text{bbox}}) + \mbox{Cov} (\mathbb{I}_{\text{class}},\mathbb{I}_{\text{bbox}}). 
\end{align*}
Under the non-negative correlation assumption $\mbox{Cov} (\mathbb{I}_{\text{class}},\mathbb{I}_{\text{bbox}})\geq 0$, the coverage bound under independence $(1-\alpha_{\text{class}})(1-\alpha_{\text{bbox}})$ remains valid, while becoming more conservative. Hence, the bound could fail only under negative association between the classification and localization errors, which we would not expect here.

Our two-step and one-step pipeline can be seen in Figure~\ref{fig:our_pipeline}, where for both the scaled and unscaled variants, the class-wise option is enabled. 

\section{Experiments}
\label{sec:experiments}
This section is organized into three parts. The first evaluates \ac{CP} applied solely to the bounding box regression task using a single class-agnostic quantile computed across all classes. The second extends this to a class-wise setting, where separate quantiles are computed per class under the assumption that the ground truth class is known. The third relaxes this assumption by incorporating \ac{CP} for the classification head via a two-step pipeline, jointly conformalizing both regression and classification. We begin by defining the evaluation metrics and experimental setup common to all three parts.

\textbf{Metrics.} We evaluate all methods on three complementary metrics. The primary metric for \ac{CP} is \textit{empirical coverage}, defined as
\begin{equation}
    \widehat{\text{Cov}} = \frac{1}{m} \sum_{i=1}^m 
    \mathbb{I}(\delta(A_i) \in \mathcal{C}(I_i)),
\end{equation}
where $m$ is the number of samples in $\mathcal{D}_{\text{eval}}$ and $\mathcal{C}(I_i)$ is the conformalized bounding box for image $I_i$ with ground truth box $\delta(A_i)$. We define coverage as complete containment of the ground truth box within the conformalized box, which preserves the standard split-CP marginal coverage guarantee (\eqref{eq:marginal_coverage}) and keeps our guarantees directly comparable to prior work~\cite{deGAdaAle2022,TimStrSak2025}. Coverage alone is insufficient to distinguish methods, since arbitrarily large intervals trivially achieve high coverage. We therefore additionally report the \textit{\ac{IoU}} between the ground truth bounding box and the outer conformalized bounding box~\citep{ZouCheKe2023}, which measures how tightly the conformalized box aligns with the ground truth. Finally, we report the \textit{interval score}~\citep{GneRaf2007}, a proper scoring rule that jointly penalizes interval width and coverage violations:
\begin{equation}
    S_{\mathrm{int},\alpha}(l,u;c) = (u - l) 
    + \frac{2}{\alpha}(l - c)\mathbb{I}\{c < l\} 
    + \frac{2}{\alpha}(c - u)\mathbb{I}\{c > u\},
\end{equation}
where $l$ and $u$ are the lower and upper bounds of the interval, $c \in \{c_{x,0}, c_{y,0}, c_{x,1}, c_{y,1}\}$ is 
the true value of the corresponding bounding box coordinate in pixel space, and $\alpha$ is the per-coordinate miscoverage level. The interval score for the full bounding box is obtained by summing over all four corners. A lower interval score indicates a better trade-off between sharpness and coverage. Together, these three metrics capture the full picture: coverage tells us whether the guarantee is met, \ac{IoU} measures alignment tightness, and the interval score penalizes both excessive width and coverage failures simultaneously.

\textbf{Statistical testing.} To assess whether differences between scaled and unscaled \ac{CP} are statistically significant, we use a two-sided paired t-test. The paired design is appropriate since both methods are evaluated on identical random splits, eliminating between-split variance. The null hypothesis is equal means between scaled and unscaled \ac{CP} for each metric.

\textbf{Experimental setup.} All experiments use 100 random splits of 80\% calibration and 20\% evaluation data. The miscoverage level $\alpha$ is specified per coordinate, such that the nominal coverage guarantee for the full bounding box is $1 - 4\alpha$ according to the Bonferroni correction. Three autonomous driving datasets are considered: KITTI~\citep{GeiLenUrt2012} (1589 predictions), BDD~\citep{YuCheWan2020} (37189 predictions), and CODA~\citep{LiCheWan2022} (4931 predictions). 
The underlying object detector is EfficientDet-d0~\citep{TanPanLe2020} trained with loss attenuation, following the setup of~\citet{KasSbeyKar2024}. Specifically, a dedicated model is trained and evaluated on KITTI, a second model is trained and evaluated on BDD, and the BDD-trained model is additionally evaluated on CODA using the eight classes common to both datasets. CODA therefore represents a cross-domain evaluation under distribution shift.

For scaled \ac{CP}, three uncertainty estimates derived from loss attenuation are evaluated: the raw uncalibrated output (LA), and two calibrated variants based on relative isotonic regression applied globally (Rel.\ \ac{IR}) or per-coordinate and per-class (Rel.\ \ac{IR} \ac{PCo} \ac{PC}). Relative isotonic regression~\citep{KasSbeyKar2023} extends standard isotonic regression by normalizing the predicted uncertainty and the residuals by the width and height of the corresponding bounding box prior to calibration, preventing the uncertainty of large objects from disproportionately influencing the calibration of smaller ones.

\subsection{Results}
\subsubsection{Class-agnostic Conformal Prediction: Regression}
\label{sec:class_agnostic}
Table~\ref{tab:results_class_agnostic} reports results for $\alpha = 0.1$ across all three datasets. All differences between scaled and unscaled \ac{CP} are statistically significant at the $1\%$ level ($p < 0.01$) for all three metrics and all datasets.

\begin{table}[!ht]
    \centering
    \caption{Results using $\alpha = 0.1$, 100 runs, and an 80/20 split for KITTI (1589 predictions), BDD (37189 predictions), and CODA (4931 predictions). $^{**}$ indicates $p < 0.01$ for the two-sided paired t-test against unscaled \ac{CP}.}
    \resizebox{\columnwidth}{!}{%
\begin{tabular}{lllllllllll}
    \toprule
    & \multicolumn{3}{c}{\textbf{KITTI}} 
    & \multicolumn{3}{c}{\textbf{BDD}} 
    & \multicolumn{3}{c}{\textbf{CODA}} \\
    \cmidrule(lr){2-4} \cmidrule(lr){5-7} \cmidrule(lr){8-10}
    Method 
    & Cov.$\uparrow$ & IoU (\%)$\uparrow$ & Int.\ Score$\downarrow$
    & Cov.$\uparrow$ & IoU (\%)$\uparrow$ & Int.\ Score$\downarrow$ 
    & Cov.$\uparrow$ & IoU (\%)$\uparrow$ & Int.\ Score$\downarrow$ \\
    \midrule
    Unscaled 
    & \textbf{0.75} & 75.01 & 130529 
    & \textbf{0.74} & 43.18 & 5154395 
    & \textbf{0.72} & 44.82 & 1731316 \\
    Scaled LA 
    & 0.72$^{**}$ & \textbf{80.40$^{**}$} & 106572$^{**}$ 
    & 0.72$^{**}$ & 46.21$^{**}$ & 4732494$^{**}$ 
    & 0.70$^{**}$ & 50.56$^{**}$ & 1539478$^{**}$ \\
    Scaled Rel.\ \ac{IR} 
    & 0.71$^{**}$ & 80.35$^{**}$ & 103879$^{**}$ 
    & 0.72$^{**}$ & 46.23$^{**}$ & 4724573$^{**}$ 
    & 0.70$^{**}$ & 50.68$^{**}$ & 1533441$^{**}$ \\
    Scaled Rel.\ \ac{IR} \ac{PCo} \ac{PC} 
    & 0.71$^{**}$ & 80.25$^{**}$ & \textbf{98261$^{**}$} 
    & 0.71$^{**}$ & \textbf{46.80$^{**}$} & \textbf{4499971$^{**}$} 
    & 0.69$^{**}$ & \textbf{51.13$^{**}$} & \textbf{1500083$^{**}$} \\
    \bottomrule
\end{tabular}
}
    \label{tab:results_class_agnostic}
\end{table}

The results reveal a consistent pattern across all datasets. Unscaled \ac{CP} achieves the highest empirical coverage, but this comes at the cost of unnecessarily wide intervals, as evidenced by its lowest \ac{IoU} and highest interval score in all settings. This behaviour is a direct consequence of the fixed quantile. A single large value is applied uniformly across all predictions regardless of their individual difficulty, inflating the conformalized boxes beyond what is needed for easy predictions while still potentially under-covering genuinely hard ones.

Scaled \ac{CP} improves substantially on both sharpness metrics while maintaining valid coverage. The absolute \ac{IoU} gain of scaled LA over unscaled \ac{CP} is 5.4\% on KITTI (75.01\% to 80.40\%), 3.0\% on BDD (43.18\% to 46.21\%), and 5.7\% on CODA (44.82\% to 50.56\%), with corresponding interval score reductions of 18.4\%, 8.2\%, and 11.1\% respectively. The largest absolute \ac{IoU} improvement is achieved by Rel.\ \ac{IR} \ac{PCo} \ac{PC} on CODA (44.82\% to 51.13\%, +6.3\%), which is also the setting with the largest domain shift. This suggests that aleatoric uncertainty scaling is particularly beneficial when the detector operates further from its training distribution, precisely the scenario where adaptive interval width matters most.

Among the scaled variants, the benefit of uncertainty calibration is modest but consistent. The uncalibrated LA already captures the largest share of the gain over unscaled \ac{CP}, while the per-coordinate and per-class calibrated variant Rel.\ \ac{IR} \ac{PCo} \ac{PC} achieves the best interval score across all datasets. This indicates that fine-grained calibration provides an additional but incremental benefit in sharpness without affecting coverage. The near-identical \ac{IoU} values of LA and Rel.\ \ac{IR} on KITTI and BDD further suggest that global isotonic regression calibration adds little over the raw uncertainty estimate when the detector is evaluated in-domain.

\begin{figure*}[!th]
    \centering
    \includegraphics[width=\columnwidth]{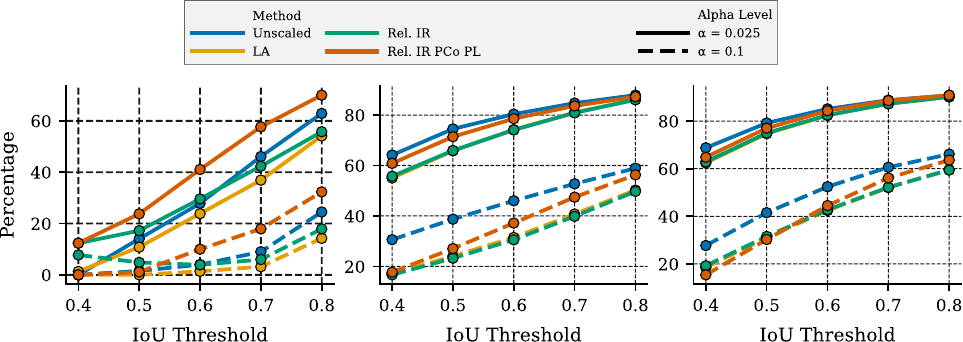}
    \caption{The percentage of ground truth boxes that the detector localized poorly, i.e, those whose \ac{IoU} with the point prediction is below a threshold, but that the conformalized box now fully contains. Shown for KITTI (left), BDD (mid), and CODA (right) at different \ac{IoU} thresholds. Colors represent \textcolor[HTML]{0072B2}{unscaled}, \textcolor[HTML]{E69F00}{LA}, \textcolor[HTML]{009E73}{Rel.\ \ac{IR}}, and \textcolor[HTML]{D55E00}{Rel.\ \ac{IR} \ac{PCo} \ac{PC}}. Solid lines (\texttt{-}) correspond to $\alpha = 0.025$ and dashed lines (\texttt{--}) to $\alpha = 0.1$.}
    \label{fig:iou_comparison}
\end{figure*}

Figure~\ref{fig:iou_comparison} shows the percentage of predictions that fall below a given \ac{IoU} threshold with the ground truth but are now fully contained within the conformalized bounding box. Two findings stand out. First, decreasing $\alpha$ increases the percentage of recovered predictions across all methods and datasets, as expected from the resulting wider conformalized boxes. Second, the recovery rates are highest on KITTI and BDD, where the detectors are evaluated in-domain, and noticeably lower on CODA, where the BDD-trained model faces distribution shift. This indicates that \ac{CP} cannot fully compensate for degraded base detector performance, and that the quality of the aleatoric uncertainty estimates from loss attenuation is tied to how well the detector generalises to the evaluation domain.

\subsubsection{Class-wise Conformal Prediction: Regression}
\label{subsection:only_class_wise_cp}

We now evaluate class-wise \ac{CP}, which computes separate calibration quantiles per ground truth class, yielding class-conditional coverage guarantees as in Equation~\eqref{eq:class_wise_cp}. Since Section~\ref{sec:class_agnostic} showed that uncalibrated LA already captures the largest gain over unscaled \ac{CP} with minimal additional cost, we restrict this evaluation to LA for the scaled variant. This evaluation assumes that the classifier is always correct, i.e.,\ ground truth class labels are used for stratification, representing a theoretical upper bound on what class-wise \ac{CP} can achieve in practice. The relaxation of this assumption is addressed in Section~\ref{sec:class_wise_cp_classification}.

\begin{table}[ht!]
    \centering
    \caption{Class-wise results for $\alpha = 0.1$, 100 runs, and an 80/20 split. Only the uncalibrated uncertainty estimate is used for scaled \ac{CP}. Coverage and \ac{IoU} are weighted by class frequency; the interval score is summed over classes and averaged over runs.}
\resizebox{\columnwidth}{!}{%
\begin{tabular}{lllllllllll}
    \toprule
    & \multicolumn{3}{c}{\textbf{KITTI}} 
    & \multicolumn{3}{c}{\textbf{BDD}} 
    & \multicolumn{3}{c}{\textbf{CODA}} \\
    \cmidrule(lr){2-4} \cmidrule(lr){5-7} \cmidrule(lr){8-10}
    Method 
    & Cov.$\uparrow$ & IoU (\%)$\uparrow$ & Int.\ Score$\downarrow$ 
    & Cov.$\uparrow$ & IoU (\%)$\uparrow$ & Int.\ Score$\downarrow$ 
    & Cov.$\uparrow$ & IoU (\%)$\uparrow$ & Int.\ Score$\downarrow$ \\
    \midrule
    Unscaled 
    & \textbf{0.74} & 75.47 & 126945 
    & \textbf{0.73} & 44.02 & 5024715 
    & \textbf{0.72} & 47.17 & 1625116 \\
    Scaled   
    & 0.71 & \textbf{80.05} & \textbf{97627} 
    & 0.72 & \textbf{46.31} & \textbf{4691174} 
    & 0.69 & \textbf{50.97} & \textbf{1432600} \\
    \bottomrule
\end{tabular}
}
    \label{tab:results_class_wise_cp}
\end{table}

The qualitative ordering from the class-agnostic setting is fully preserved. Scaled \ac{CP} consistently achieves a higher \ac{IoU} and lower interval score than unscaled \ac{CP} across all three datasets, while unscaled \ac{CP} retains higher empirical coverage. The key finding from comparing Table~\ref{tab:results_class_wise_cp} with Table~\ref{tab:results_class_agnostic} is that class-wise stratification reduces the interval score for both methods without meaningfully changing coverage or \ac{IoU}. On KITTI, the interval score decreases by 2.7\% for unscaled (130528 to 126944) and by 8.4\% for scaled (106571 to 97627). The larger reduction for scaled \ac{CP} suggests that class-conditional quantiles interact more favorably with the scaling by aleatoric uncertainty. When calibration is stratified by class, the per-class uncertainty estimates from loss attenuation are better aligned with the difficulty distribution of each class, yielding sharper intervals without additional coverage loss. The improvement is more pronounced on CODA than on BDD, consistent with the greater heterogeneity of object appearances under domain shift, where class-specific calibration of the uncertainty estimates provides more benefit. We note that the weighted averages in Table~\ref{tab:results_class_wise_cp} reflect class imbalance across all three datasets. For minority classes with few calibration samples, class-wise quantiles may be less stable~\citep{AngBat2021}, which partially explains why the gains over the class-agnostic setting are moderate rather than large. This is further discussed in Section~\ref{sec:limitations}.

\subsubsection{Joint Conformal Prediction: Classification and Regression}
\label{sec:class_wise_cp_classification}

The class-wise results in Section~\ref{subsection:only_class_wise_cp} rely on access to the true class labels at inference time, which is unavailable in practice. We now address this by incorporating \ac{CP} for the classification head via a two-step pipeline. A prediction set for the class is first constructed using RAPS, and the conformalized bounding box is then conditioned on the predicted class set by selecting the largest class-specific quantile. Under the non-negative correlation assumption between classification and regression \ac{CP}, this yields a joint coverage of at least $(1-\alpha_{\text{class}})(1-\alpha_{\text{bbox}})$ as shown in Equation~\eqref{eq:cp_for_class_and_bbox_coverage}.

This evaluation is restricted to KITTI, as the two-step pipeline requires the classifier to be sufficiently certain to produce small, informative prediction sets. On BDD the class distribution is more imbalanced, and on CODA the model operates under distribution shift from its BDD training domain; both factors increase classification uncertainty, leading to larger prediction sets and overly conservative bounding box intervals. We compare unscaled and scaled \ac{CP} with uncalibrated LA only.

\textbf{Selecting the classification CP algorithm.}
Table~\ref{tab:results_cp_only_for_classes} compares APS and RAPS for $\alpha_{\text{class}} = 0.01$ on KITTI. All methods satisfy the coverage guarantee but they differ substantially in the size of prediction sets. APS produces sets roughly twice as large as both RAPS variants (3.457 vs.\ 1.762 and 1.765), which would directly inflate the bounding box intervals in the two-step pipeline by propagating larger worst-case quantiles. Between the two RAPS variants, allowing empty sets reduces both coverage (99.50\% vs.\ 99.85\%) and set size marginally. Empty prediction sets are inadmissible in our pipeline since they yield a class quantile of zero, collapsing the bounding box interval to a degenerate estimate. We therefore use RAPS without empty sets for all subsequent experiments.

\begin{table}[ht!]
    \centering
    \caption{Coverage and mean prediction set size for APS and RAPS variants evaluated over 100 random splits with $\alpha_{\text{class}} = 0.01$ on KITTI.}
\begin{tabular}{lll}
    \toprule
    Method & Coverage$\uparrow$ & Mean Set Size$\downarrow$ \\
    \midrule
    APS & \textbf{99.94}\% & 3.46 \\
    RAPS with empty sets & 99.50\% & \textbf{1.76} \\
    RAPS w.o.\ empty sets & 99.85\% & 1.77 \\
    \bottomrule
\end{tabular}
    \label{tab:results_cp_only_for_classes}
\end{table}

\begin{figure*}[t!]
    \centering
    \begin{minipage}{0.3\textwidth}
        \centering
        \includegraphics[width=\textwidth]{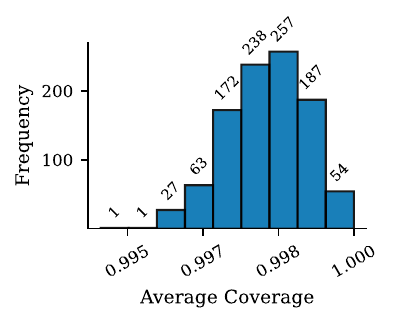}
        \\(a) $\alpha = 0.01$
    \end{minipage}
    \begin{minipage}{0.3\textwidth}
        \centering
        \includegraphics[width=\textwidth]{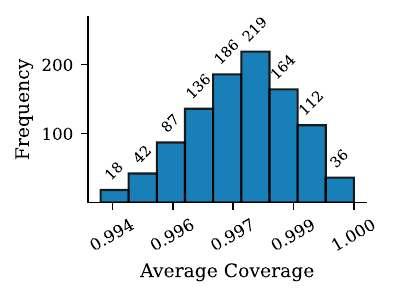}
        \\(b) $\alpha = 0.05$
    \end{minipage}
    \begin{minipage}{0.3\textwidth}
        \centering
        \includegraphics[width=\textwidth]{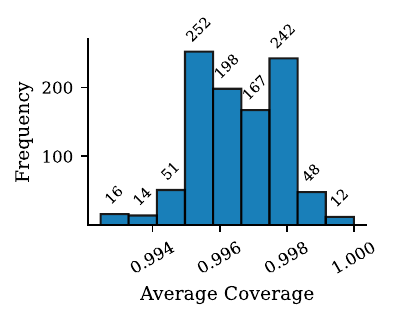}
        \\(c) $\alpha = 0.1$
    \end{minipage}
    \\[1em]
    \begin{minipage}{0.3\textwidth}
        \centering
        \includegraphics[width=\textwidth]{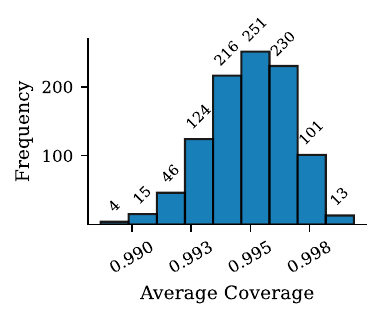}
        \\(d) $\alpha = 0.01$
    \end{minipage}
    \begin{minipage}{0.3\textwidth}
        \centering
        \includegraphics[width=\textwidth]{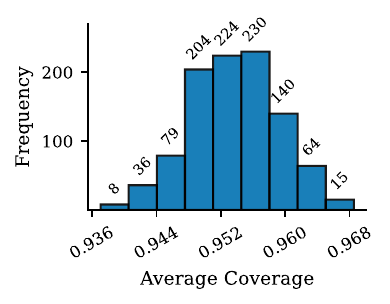}
        \\(e) $\alpha = 0.05$
    \end{minipage}
    \begin{minipage}{0.3\textwidth}
        \centering
        \includegraphics[width=\textwidth]{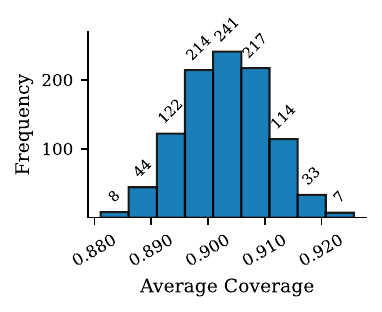}
        \\(f) $\alpha = 0.1$
    \end{minipage}
    \caption{Distribution of empirical average coverage over 1000 runs using 1589 evaluation samples on KITTI. The top row shows RAPS without empty prediction sets and the bottom row shows RAPS with empty prediction sets, evaluated at $\alpha \in \{0.01, 0.05, 0.1\}$.}
    \label{fig:historgrams_for_adaptivty_avg_cov}
\end{figure*}

Figure~\ref{fig:historgrams_for_adaptivty_avg_cov} shows the distribution of empirical coverage over 1000 runs for both RAPS variants at three values of $\alpha$. For RAPS without empty sets, the coverage distribution is sharply concentrated near 1.0 regardless of $\alpha$. This is a consequence of EfficientDet's high classification accuracy on KITTI (above 99\%). The conformal quantile $\hat{q}$ is already large enough to include the true class with near-certainty for any reasonable $\alpha$, such that additional regularization with RAPS does only marginally provide benefits. For RAPS with empty sets, the distribution correctly shifts toward $1-\alpha$ as $\alpha$ increases, confirming that the \ac{CP} guarantee is properly calibrated when empty sets are permitted. The near-perfect coverage of RAPS without empty sets is therefore a property of the in-domain detector accuracy on KITTI rather than a failure of the \ac{CP} procedure itself.

Table~\ref{tab:results_cp_for_bboxes_and_classification} shows results for the two-step pipeline at two miscoverage levels. As expected, tightening the requirement from $\alpha = 0.1$ to $\alpha = 0.05$ increases empirical coverage for both methods while reducing \ac{IoU} and increasing the interval score, reflecting the wider conformalized boxes required to meet the stricter guarantee.

\begin{table}[ht!]
    \centering
    \caption{Results for the two-step pipeline on KITTI using $\alpha_{\text{class}} = 0.01$ and $\alpha \in \{0.05, 0.1\}$ per bounding box corner. 100 runs, 80/20 split, 1589 evaluation samples.}
\begin{tabular}{lllllll}
    \toprule
    & \multicolumn{3}{c}{\textbf{$\alpha = 0.1$}} 
    & \multicolumn{3}{c}{\textbf{$\alpha = 0.05$}} \\
    \cmidrule(lr){2-4} \cmidrule(lr){5-7}
    Method 
    & Cov.$\uparrow$ & IoU (\%)$\uparrow$ & Int.\ Score$\downarrow$ 
    & Cov.$\uparrow$ & IoU (\%)$\uparrow$ & Int.\ Score$\downarrow$  \\
    \midrule
    Unscaled 
    & \textbf{0.77} & 68.48 & 147854 
    & \textbf{0.88} & 59.03 & 267126 \\
    Scaled   
    & 0.76 & \textbf{76.65} & \textbf{103020} 
    & \textbf{0.88} & \textbf{70.19} & \textbf{162895} \\
    \bottomrule
\end{tabular}
    \label{tab:results_cp_for_bboxes_and_classification}
\end{table}

In comparison with the class-agnostic regression results in Table~\ref{tab:results_class_agnostic} for $\alpha = 0.1$, three observations emerge. First, incorporating classification \ac{CP} increases empirical coverage for both methods (unscaled: 0.75 to 0.77; scaled: 0.72 to 0.76). This is expected: the two-step pipeline selects the largest quantile among the predicted class set, which is on average more conservative than a single class-agnostic quantile. Second, \ac{IoU} decreases for both methods (unscaled: 75.01 to 68.48; scaled: 80.40 to 76.65), reflecting this more conservative interval construction. Third, and most importantly, the interval score decreases for scaled \ac{CP} (106571 to 103020, $-3.3\%$) but increases substantially for unscaled \ac{CP} (130528 to 147854, $+13.3\%$). This divergence is the key result of the two-step evaluation. For scaled \ac{CP}, conditioning on the predicted class set and scaling by aleatoric uncertainty jointly produce intervals that remain well-calibrated to the difficulty of individual predictions even under the more conservative class-conditional quantile selection. For unscaled \ac{CP}, the fixed-width intervals are further inflated by the worst-case class quantile with no compensating mechanism, substantially worsening the sharpness-coverage trade-off. This finding provides the strongest argument for scaled \ac{CP} in the full two-step pipeline. Not only does it produce sharper intervals in isolation, it degrades more gracefully when combined with classification uncertainty.

\section{Conclusion}
\label{sec:conclusion}
We systematically compare unscaled and scaled \ac{CP} for multi-class object detection on three autonomous driving datasets, including an in-domain and a cross-domain evaluation under distribution shift. Scaled \ac{CP}, where prediction intervals are adapted using aleatoric uncertainty estimates derived from loss attenuation, consistently produces sharper and better-aligned conformalized bounding boxes than unscaled \ac{CP}, as measured by the \ac{IoU} and interval score. Crucially, this improvement in sharpness does not negatively affect coverage guarantee. Both variants maintain valid marginal coverage, with unscaled \ac{CP} tending to over-cover beyond the nominal level.

The largest performance gain is achieved by switching from unscaled to scaled \ac{CP} with an uncalibrated uncertainty estimate. Further calibrating the uncertainty via relative isotonic regression yields only marginal additional improvements, suggesting that the raw aleatoric uncertainty from loss attenuation already captures the most relevant variation in prediction difficulty. Class-wise \ac{CP} reduces the interval score for both variants with negligible effect on coverage, and with the reduction being more pronounced for scaled \ac{CP}. This suggests that class-conditional quantiles interact more favourably with aleatoric uncertainty scaling, yielding sharper intervals when prediction difficulty varies across classes.

For the two-step pipeline combining classification and regression \ac{CP}, scaled \ac{CP} again outperforms unscaled \ac{CP}, and notably its interval score decreases relative to the regression-only setting while that of unscaled \ac{CP} increases. This reveals an important asymmetry: scaled \ac{CP} degrades more gracefully when combined with classification uncertainty, making it the preferred choice for the full two-step pipeline. This pipeline further requires that the classification \ac{CP} algorithm produces non-empty prediction sets, making RAPS without empty sets the appropriate choice, and benefits from high base classifier accuracy.

Taken together, our results demonstrate that scaling \ac{CP} intervals by aleatoric uncertainty estimates is a simple, computationally efficient, and principled way to improve the sharpness of conformalized object detectors without sacrificing validity. Moreover, this finding is robust across datasets, uncertainty calibration strategies, and pipeline configurations.

\section{Limitations}
\label{sec:limitations}
A first limitation concerns the size of the calibration set in the class-wise setting.~\citet{AngBat2021} recommend approximately 1000 calibration samples per class for stable \ac{CP} quantile estimates. This requirement is not met for all classes in our datasets due to class imbalance, which may reduce the stability of the class-wise quantiles and partially explains why the gains from class-wise \ac{CP} over the class-agnostic setting are moderate. Future work could investigate the effect of enforcing a minimum per-class calibration set size, for example, through oversampling or dataset curation.

Second, the hyperparameters of RAPS are not tuned in this work and are set to the values suggested by~\citet{AngBat2021}. Tuning them could further reduce prediction set sizes and improve the sharpness of the conformalized bounding boxes in the two-step pipeline.

Third, the theoretical guarantees of \ac{CP} rely on the exchangeability of calibration and test data. This assumption may be violated in real-world autonomous driving scenarios, for example, under changes in weather, lighting, or sensor conditions between calibration and deployment. The cross-domain evaluation on CODA, where the BDD-trained model is evaluated under distribution shift, already illustrates the practical consequences of this violation. Recovery rates are lower and the benefits of uncertainty scaling are less pronounced than in the in-domain settings. Developing \ac{CP} variants that are robust to distribution shift, for example, through weighted conformal prediction~\citep{TibFoyBarCan2019}, represents a promising direction for future work.

Fourth, the two-step classification--regression pipeline has two conceptual limitations. Its joint coverage guarantee assumes non-negative correlation between the classification and localization conformal procedures; as discussed in Section~\ref{sec:joint_cp}, classification and localization difficulty are expected to be positively associated in practice, which renders the product bound conservative rather than invalid, although an explicit empirical characterization of this dependence would strengthen the justification. In addition, the class prediction set is constructed from the classification scores of the single box selected by the detector, and therefore does not account for the fact that different plausible boxes could imply different class probabilities. A genuinely joint conformal set defined over the four box coordinates and the object class would address both points and is a natural next step.

Finally, two methodological extensions are worth noting. Instead of the conservative Bonferroni correction, the four coordinate-wise errors could be aggregated with a dependence-aware correction such as the max-rank procedure of~\citet{TimStrSak2024}, which exploits the positive dependence between coordinate errors to yield tighter box-level intervals. Moreover, because the aleatoric uncertainty estimates act as a general difficulty signal rather than a box-specific quantity, the same scaling could be applied to richer geometric representations such as contours, polygons, or segmentation masks, with complete containment relaxed to task-specific coverage notions such as occupied-area coverage via Conformal Risk Control~\citep{AngBatFis2024}.

\acks{This work has been partially funded by the Pilot Program for Core Informatics (KiKIT) at the KIT of the Helmholtz Association. The authors also acknowledge support by the state of Baden-Württemberg through bwHPC.}

\bibliography{pmlr-sample}
\end{document}